# A Revisit of Fake News Dataset with Augmented Fact-checking by ChatGPT

**Zizhong Li    Haopeng Zhang    Jiawei Zhang**
University of California, Davis
{zzoli, hapzhang, jiwzhang}@ucdavis.edu

## Abstract

The proliferation of fake news has emerged as a critical issue in recent years, requiring significant efforts to detect it. However, the existing fake news detection datasets are sourced from human journalists, which are likely to have inherent bias limitations due to the highly subjective nature of this task. In this paper, we revisit the existing fake news dataset verified by human journalists with augmented fact-checking by large language models (ChatGPT), and we name the augmented fake news dataset ChatGPT-FC. We quantitatively analyze the distinctions and resemblances between human journalists and LLM in assessing news subject credibility, news creator credibility, time-sensitive, and political framing. Our findings highlight LLM's potential to serve as a preliminary screening method, offering a promising avenue to mitigate the inherent biases of human journalists and enhance fake news detection.

## 1 Introduction

Fake news, or misinformation, has a long history in societies (Conroy et al., 2015; Burkhardt, 2017; Lazer et al., 2018). In recent years, social networks have provided a fertile ground for breeding and spreading fake news, creating great confusion and misunderstanding about important social and political issues, especially for people with little access to factual information sources. For example, a study shows that many high school students can not distinguish fake news sources on Facebook (Wineburg and McGrew, 2016). Moreover, unlike some traditional suspicious information like spam (Dada et al., 2019), fake news is not only challenging to identify but also has significant social implications, as seen in the Pizzagate shooting incident in 2016 (Kang and Goldman, 2016). In such circumstances, considerable efforts have been devoted to detecting fake news (Qi et al., 2022; Si et al., 2022) and mitigating its negative impact on society.

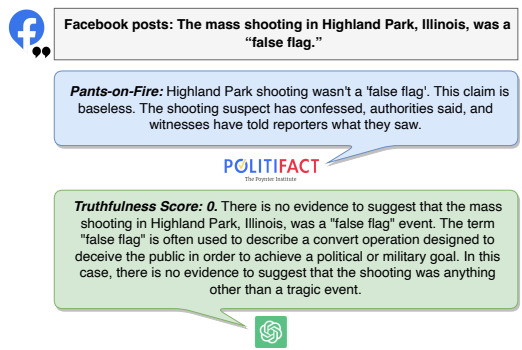

Figure 1: A comparison between the website journalists' justification and ChatGPT's response for the same potential fake statement, showing that ChatGPT's generated response could provide additional information for fact-checking from another perspective. (Due to the limited space, we provide the summarized content from the website's fake-checking report article.)

However, the current fake news detection methods still heavily rely on human fact-checking resources (D'Ulizia et al., 2021) such as social media accounts and authoritative websites like *PoliticFact* (Zhang et al., 2020; Shu et al., 2017) and *Channel4 News* (Vlachos and Riedel, 2014). Although these manual verification articles have shown their effectiveness, they may still have some shortcomings when being used for training machine learning-based fact-checking methods. One of the most important issues is that the human-labeled data may have their inherent biases due to individual subjectivity and preference (Van der Linden et al., 2020; Grady et al., 2021), since websites like *PoliticFact* also have their funding source and this may affect their reports and stance.

In the past few months, we have observed significant progress on LLMs (large language models) (Devlin et al., 2018; Floridi and Chiriatti, 2020) and pre-training techniques (Rethmeier and Augenstein, 2023; Li et al., 2021). The superior performance of LLMs like ChatGPT (OpenAI, 2023; Ouyang et al., 2022), especially in the misinformation iden-tification task (Bang et al., 2023), also reveals that it is likely able to make a significant contribution to the fake news detection task and may potentially mitigate the issues that currently exist in the manual fake-checking data. More specifically, ChatGPT provides an opportunity to study and analyze fake news from a different perspective instead, like Figure 1 shows. Pre-trained with the massive data across the Internet and document sources, ChatGPT is capable to access more comprehensive information sources when generating the fact-checking for potential fake news.

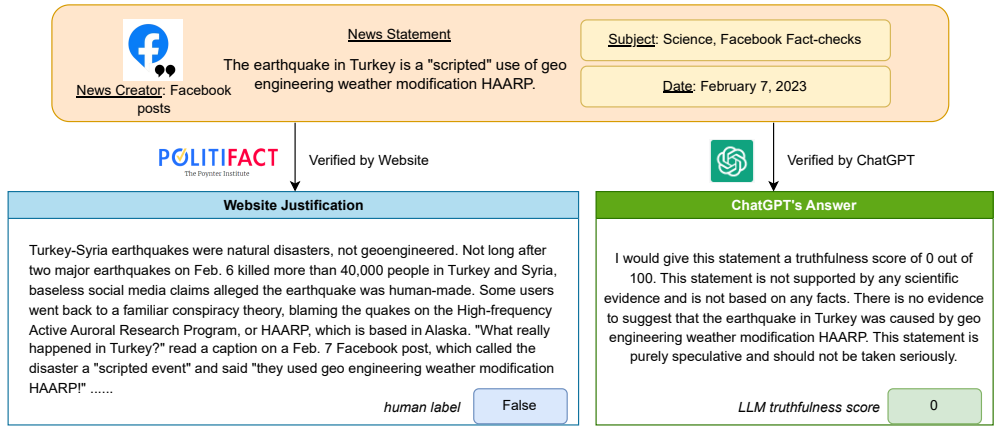

Figure 2: An Example of ChatGPT-FC Dataset.

Aiming to demonstrate that ChatGPT can provide complementary information which is different from human journalists, we propose revisiting the existing fake news dataset by incorporating an augmented fact-checking method using ChatGPT. Formally, we name this enhanced dataset as ChatGPT-FC. As illustrated in Figure 2, ChatGPT-FC consists of three main information components: the news statement and its corresponding attributes (i.e., subject, news creator, date), the fact-checking report written by human journalists, and fact-checking report generated by ChatGPT. Both news statements and human-written fact-checking reports are sourced from the *PoliticFact* website[1], including about 22,000 news statements covering political and social issues from 2007 to 2023. Simulating the process of human journalist fact-checking, we design a fact-checking prompt for ChatGPT to generate corresponding responses, which serve as the LLM component within the ChatGPT-FC dataset. We first conducted comprehensive data analysis, focusing on the aspects of *subject credibility*, *news creator credibility*, and *time-sensitive*. Furthermore, we introduce a framework to analyze the inherent biases present in ChatGPT-FC from both the perspective of textual content and assessment criteria. Our findings demonstrate that human journalists' fact-checking tends to be more stringent than LLM. However, the evaluation criteria used by human journalists are more subjective and less quantifiable than LLM and are particularly susceptible to political framing biases. In contrast, LLM-based fact-checking relies on factual evidence and is more objective in nature.

In summary, our main contributions in this paper are summarized as follows:

- We introduce the ChatGPT-FC dataset, which includes the fact-checking reports written by both human journalists and the complementary information generated by ChatGPT. To the best of our knowledge, ChatGPT-FC is the first dataset that provides not only the human-verified fact-checking source but also another qualified checking view from LLMs.

- We developed an evaluation framework to assess the quality of fact-checking content from both human journalists and LLMs, showing that the involvement of LLMs in fact-checking provides a valuable perspective, complementing the subjective nature of human ratings influenced by various factors.

## 2 Related Work

### 2.1 Fake News Dataset

As one of the most crucial components of fake news detection, previous researchers have put in a great deal of effort in obtaining datasets. From the first release of a fake news detection dataset (Vlachos and Riedel, 2014) to nowadays, over 25 datasets covering a range of news domains including society (Torabi Asr and Taboada, 2019; Papadopoulou et al., 2019), politics (Barrón-Cedeno et al., 2019; Wang, 2017), and technology (Barbado et al., 2019; Tacchini et al., 2017) have been utilized in various fake detection and fact-checking models. Moreover, during COVID-19, many rumors and false news about the pandemic have emerged and spread on the internet, leading to the creation of several benchmark datasets related to the issue (Patwa et al., 2021; Yang et al., 2021; Kim et al., 2021). However, almost all of the datasets are sourced from mainstream media or from social media such as Facebook and Twitter. In other words, the existing fake news datasets are sourced from manual verification, which does not provide a multi-view for the news and may have some inherent biases.

---
[1] https://www.politifact.com/

Table 1: The ChatGPT-FC Dataset Statistics

|  | Overall Statistics | |
| --- | --- | --- |
| # Total Statements | 22,337 | |
| Time span | 11/15/2007 - 02/15/2023 | |
| # Statement Subjects | 154 | |
| # Statement Creators | 4,636 | |
| # Statement Creators' Affiliations | 23 | |
| # Fact-checking Journalists | 577 | |
|  | Human Journalists | ChatGPT Answer |
| Judging Criteria | 6-level label | truthfulness score (0~100) |
| # True Statements | 7,016 | 14,963 |
| # Fake Statements | 15,321 | 7,364 |
| Valid Label/Score Percentage | 100% | 99.96% |
| Average Words Per Fact-checking Article | 949.46 | 65.24 |
| Average Sentences Per Fact-checking Article | 27.65 | 3.32 |

## 2.2 Bias in Dataset

For news media, previous efforts have revealed and analyzed the biases in terms of political leaning (Budak et al., 2016; Lewandowsky et al., 2020), gender (Asr et al., 2021; Rao and Taboada, 2021), and race (Scharrer and Ramasubramanian, 2015), which shows that the presence of biases in datasets with manual annotations is a common occurrence. Unfortunately, for machine learning-based models, the human-labeled dataset will incorporate and propagate its inherent human and societal biases into its results (Raza et al., 2022; Guo et al., 2022) and thus will affect the performance and fairness of the model (Schuster et al., 2019). For example, the recruitment model trained by Amazon tended to discriminate against female applicants due to data bias (Dastin, 2018). Therefore, alleviating or eliminating biases in datasets and making the model fairer is a very important task.

## 3 ChatGPT-FC Dataset

The data we collected and studied in the ChatGPT-FC dataset comes from the news fact-checking website *PolitiFact*[2], spanning from 2007 to 2023. *PolitiFact* was founded in 2007 as an election-year project of the Tampa Bay Times, where journalists fact-check influential statements made by politicians or on social media. We crawled a total number of 22,337 news statements from 4,636 creators across 23 affiliations in the ChatGPT-FC dataset. Each fact-checking report provides a clear fact-checking label selected from the pre-defined 6-level label set ({*True*, *Mostly True*, *Half True*, *Barely True*, *False*, *Pants on Fire*}), where *True* represents the completely accurate statements and *Pants on Fire* represents the completely false statements based on the news authenticity. Among all the statements checked by journalists, there are 7,016 true statements, and 15,321 fake statements[3].

On the other side, to utilize the ChatGPT API (text-davinci-003) for fact-checking the crawled news statements, we create a fact-checking prompt that includes the raw statement and fact-checking instructions. For instance, based on the statement depicted in Figure 2, we send the following query to ChatGPT API, as the provided examples in Appendix A shows. This query prompt will guide ChatGPT to generate truthfulness scores and supporting evidence, enabling us to evaluate the quality of ChatGPT's generated text in terms of its semantic content and fact-checking score.

Given the designed prompt, we collect 22,337 fact-checking answers from ChatGPT, which form another part of our dataset. Among all of the news statements' results checked by ChatGPT, there are 14,961 true statements (whose score is equal to or over 50) and 7,359 fake statements (whose score is below 50). More detailed statistics of our dataset are summarized in Table 1.

---
[2] https://www.politifact.com/
[3] True statement means the fact-checking label belongs to *True*, *Mostly True* or *Half True*, and false statement means the fact-checking label belongs to *Barely True*, *False* or *Pants on Fire*.

Compared with the fact-checking reports written by human journalists, the reports generated by ChatGPT are much more concise since it is designed as a chatbot. Specifically, the average word count and sentence count for ChatGPT-generated fact-checking articles is 65.24 and 3.32 respectively, compared to 949.46 and 27.65 for human journalists. Additionally, as shown in the statistical results in Table 1, there are still some instances where ChatGPT could not provide a truthfulness score (with a percentage of 0.04). Appendix B presents all kinds of cases when ChatGPT does not show a valid score.

## 4 Basic Quantitative Analysis

### 4.1 Subject Credibility Analysis

The subject refers to the topics of the news statement and the corresponding fact-checking articles in this paper. In our ChatGPT-FC dataset, there exist 154 subjects about various social and political issues. Here we provide the news credibility distribution on 7 subjects with the largest number of news statements in the dataset as Figure 3 shows. The statement credibility labels come from both human journalists and ChatGPT, respectively.

According to Figure 3, it can be seen that if the topic concerns healthcare and elections or the news is propagated via social media, the website fact-checking results will likely incline toward the fake group. Specifically, the subject *Facebook Fact-checking* covers the largest number of articles, which is 3,960. Among these articles, 216 (5.45%) of them are true and 3,744 (94.55%) of them are false. The second largest subject is *Health Care* with 1,448 articles, of which 677 (46.75%) are true and 771 (53.25%) are false. However, articles which belong to the economy and taxes subjects are likely to be grouped into the trusted side on the website. For example, the third largest subject *Economy* has 1,397 articles, of which 846 (60.56%) are true and 551 (39.44%) are false.

Different from the fact-checking results shown on the website, the results given by ChatGPT tend to suggest higher credibility for the news statements under the main subjects. That is, among the valid 3,957 answers under the subject *Facebook Fact-checking*, 1,164 (29.42%) of its answers receive scores for true statements and 2,793 (70.58%) receive scores for false statements; among the valid 1,447 answers under the subject *Health Care*, 1,038 (71.73%) of its answers receive scores for true statements and 409 (28.27%) receive scores for false statement; and among the valid 1,447 answers under the subject *Economy*, 1,082 (74.78%) of its answers receive scores for true statements and 315 (21.77%) receive scores for false statement. Therefore, it is also clear that ChatGPT's authenticity tendency in most subjects is consistent with that given by the website, but in subjects like *Health Care*, ChatGPT provides different judgment tendencies.

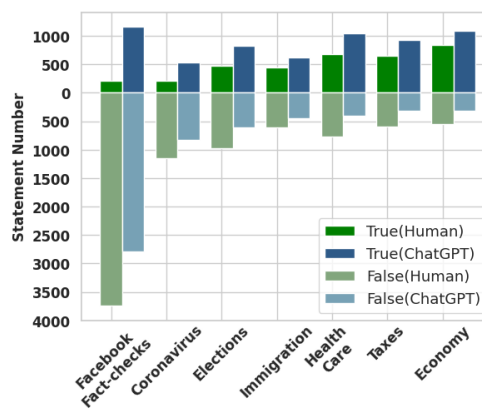

Figure 3: Subject Credibility Statistic

### 4.2 News Creator Credibility Analysis

The news creator denotes the person who made the news statements in this paper. In Figure 4, we provide statistics on credibility distribution for important politicians and social media accounts impacting politics and societies greatly. For news fact-checking articles, most social media statements are evaluated as false, comprising over 90% of all statements. The most notable creator is "Viral image", where 1035 out of total 1082 statements (i.e., 95.66%) are considered to be false by human journalists. For the news statements from Democrats, the journalists tend to label their statements to be true. For instance, for the statement from "Barack Obama", 317 out of the total 595 statements (i.e., 53.28%) are labeled to be true. Meanwhile, from the news statements from the Republicans, like "Donald Trump", the ratio of the true statements is much lower, where 203 out of total 967 statements are labeled to be true (i.e., 21.0%).

Distinct from the human journalists, the ChatGPT model presents very different credibility label distributions for these statements. For both social media accounts and specific politicians, it gives a much higher credibility score compared with human journalists. Specifically, from our case studies, ChatGPT answers improve the credibility of statements from Democratic politicians by 38.49% to 41.0%, Republican politicians by 24.40% to 47.39%, and social media accounts by 21.14% to 38.78%.

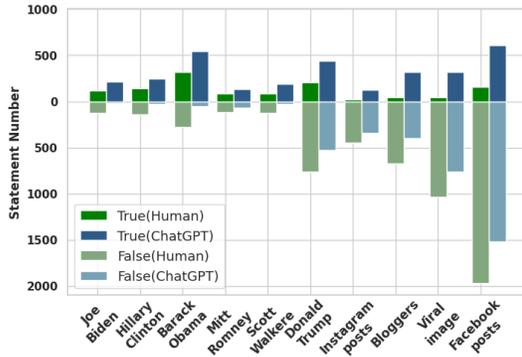

Figure 4: News Creator Credibility Statistic

### 4.3 Time-sensitive Analysis

The performance of LLMs relies heavily on their training corpus (Qiu et al., 2020) quality, particularly the timeliness of knowledge within the corpus, which significantly impacts the model's performance. According to OpenAI's information[4], ChatGPT (text-davinci-003) is trained on a corpus that is current until June 2021. It is important to investigate whether ChatGPT can maintain accurate fact-checking for news statements that fall outside the scope of its training corpus.

In Figure 5, we provide statistics of ChatGPT's fact-checking results during and after the training corpus time range. Here, we divide the truthfulness score ChatGPT gives into 5 ranges, from 100~80 to 20~0. The statistical result shows that ChatGPT gives a higher proportion of *True* or *Mostly True* judgments (i.e., the proportion of scores in the range of 100~80 is 58.28%) compared to the condition after June 2021 (i.e., the proportion of scores in the range of 100~80 is 53.18%). At the same time, in recent years, the *PolitiFact* website also includes suspicious statements from online social media for fact-checking, most of which are labeled to be *False* by human journalists. Therefore, from

---
[4]https://platform.openai.com/docs/models/gpt-4

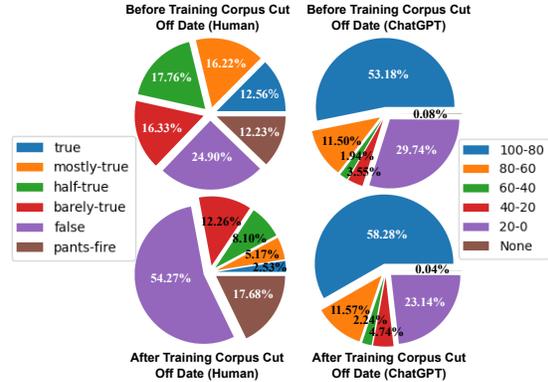

Figure 5: Label (left) and Truthfulness Score (right) Distribution at Different Time Periods

Figure 5, we observe that human journalists classify more of the news after the training corpus cutoff date as fake news, i.e., 54.27% proportion of *False* and 17.68% proportion of *Pants on Fire*, whose label distribution is significantly different compared with before.

Therefore, due to a lack of up-to-date world knowledge, ChatGPT has a lower capability in distinguishing inaccurate information for news after its training corpus cut-off date. Moreover, the increasingly spreading of false current affairs news on social media in recent years could also worsen this phenomenon, resulting in a more lenient judgment of ChatGPT.

### 4.4 Opinion Consistency Analysis

Based on the previous analysis, it is evident that ChatGPT exhibits lenient and positive judgments compared to human journalists in terms of subjects, news creators, and up-to-date news. Additionally, the general opinion consistency between human journalists and ChatGPT in our dataset warrants further examination. Here, we provide the statistical result of ChatGPT score distribution over each label in Figure 6. As introduced before, we divide the truthfulness score from ChatGPT into 5 ranges, and we further zoom into each range to investigate the article labels given by human journalists. For instance, according to the plot, for the 11,887 news articles with ChatGPT truthfulness score in the range 100~80, about 6,869 news articles are true and 5,018 news articles are false as labeled by human journalists. However, when it comes to the lower credibility news, these two sides' opinions are likely to be consistent: for the news articles with ChatGPT truthfulness score in the 20~0 range, the majority of them are also labeled as false by human journalists. It indicates that language models can effectively pre-filter potential fake news articles for human journalists.

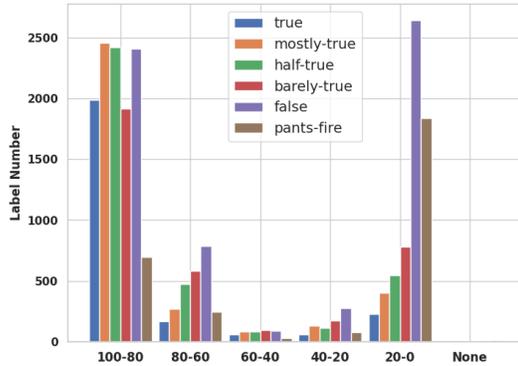

Figure 6: ChatGPT Truthfulness Score Distribution Over Human Journalist Label

## 5 Inherent Bias Analysis

Besides the above news credibility distribution analysis, we also investigate the widely existing bias issues in fake news detection. Bias can be seen as "unfairness" in the source data because of the human annotator's inherent bias and the integration of prejudiced assumptions (i.e., maybe shown in gender, race, demographics, economic status, religion, etc.) into LLMs during the training stage. Previous studies show that news organizations express their ideological bias not by directly advocating for a preferred political party, but rather by disproportionately criticizing one side (Budak et al., 2016). With a deep investigation into this phenomenon, we quantitatively analyze the inherent bias by taking the website's selection of news from three major speaker affiliations (i.e., Social Media, Democrat, and Republican) as a starting point. We categorize the bias type into *content bias* and *credibility assessment criteria bias* based on their forms. The former refers to biases present in the textual content (i.e., the discriminatory language), while the latter refers to biases present in the classification results (i.e., the distribution of "true" and "false").

### 5.1 Content Bias

We apply a frequency-based approach to quantitatively analyze the potential biases in the fact-checking content present in the dataset. Specifically, we calculate the occurrence of words from the linguistic bias lexicons (Riloff and Wiebe, 2003) within the content of both sides (i.e., journalists and ChatGPT). We focus on the frequency of strongly subjective words and weakly subjective words appearing in the fact-checking content. That is, a strongly subjective clue refers to a term that rarely carries a meaning devoid of subjectively, such as *anti-*, *deplete*, and *entreat*; and a weakly subjective clue is a term that commonly has both subjective and objective uses, like *accept*, *full*, and *patient*. By calculating the average occurrence of the strongly and weakly subjective words in each sentence, we are able to assess the degree of the subjective biases within the content. To explore potential biases related to three major speaker affiliations, we focused on fact-checking content specifically associated with these entities.

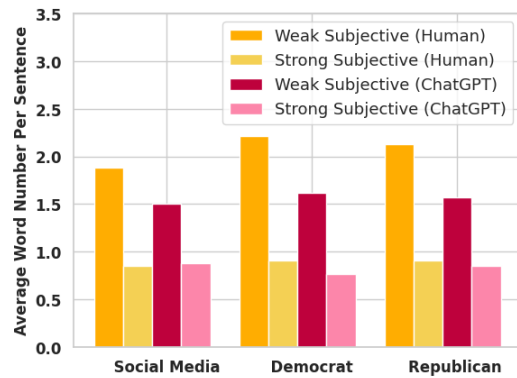

Figure 7: ChatGPT-FC Content Inherent Bias Analysis

Based on the results of our analysis from Figure 7, it is clear that both human journalists and ChatGPT-generated content exhibit a higher frequency of weakly subjective words compared to strongly subjective words. Furthermore, the frequency of both weakly and strongly subjective words in ChatGPT-generated fact-checking content is significantly lower than that in the content written by human journalists for the three major organizations. Moreover, we observe that while the frequency of strongly subjective words is similar between human journalists and ChatGPT content across these three speaker affiliations, the difference lies in the aspect of weakly subjective words. Specifically, weakly subjective words tend to appear more frequently in fact-checking articles related to the two political parties, namely *Democrat* (i.e., 2.21) and *Republican* (i.e., 2.12), compared to *Social Media* (i.e., 1.88) on the human journalists' side in ChatGPT-FC. However, the likelihood of weak subjective words appearing remains consistent across the three major speaker affiliations on the ChatGPT side.

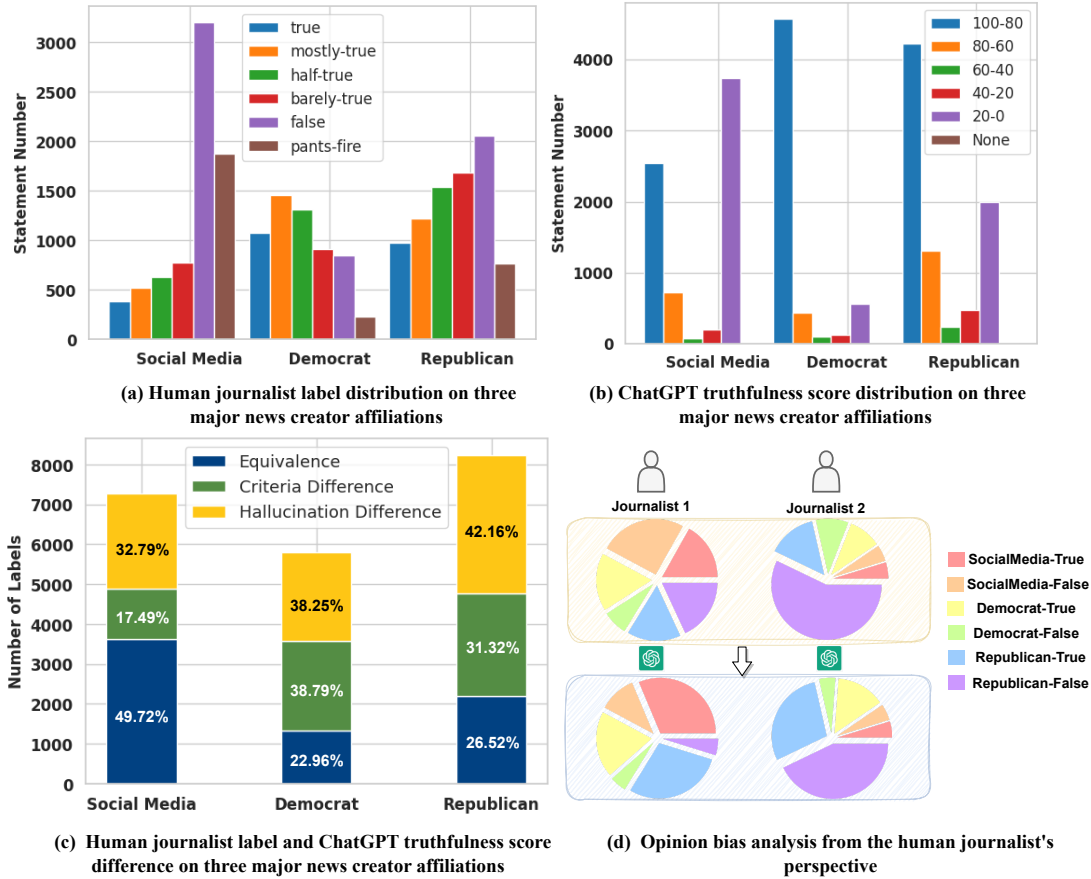

Figure 8: ChatGPT-FC Dataset Credibility Assessment Criteria Bias Analysis

## 5.2 Credibility Assessment Criteria Bias

For credibility assessment criteria bias, it is essential to analyze the difference between human journalists and ChatGPT in their labeling (or scoring) of these three major speaker affiliations. Figure 8 (a) and Figure 8 (b) show the label and score distribution on these three main afflictions. It is evident that human journalists at the *PolitiFact* tends to choose and then discredit the erroneous statements of *Republicans* and give more positive evaluations of *Democrats* in its news selection, verifying the views from previous research (Budak et al., 2016).

In addition, we also analyze the differences in the labeling/scoring criteria for the news belonging to three major speaker affiliations between human journalists and ChatGPT response. Following the scoring criteria provided by *Politifact* [5], we establish an equivalence and difference relationship between the labels and truthfulness scores, as shown in Table 2. On the other hand, when the relationship between labels and truthfulness scores is not equivalent, the difference between them should not be overlooked (i.e., when the label is *True*, there is a significant distinction between a score of 80 and a score of 0). Therefore, we categorize the differences into two types: *Criteria Difference* and *Hallucination Difference*. The former captures the overall consensus of label and truthfulness score alignment with the fact-checking results (i.e., label is in true set and score is equal or over 50, or label is in false set and score is below 50), while the latter accounts for the overall disagreement between label and truthfulness score (i.e., label is in true set and score is below 50, or label is false set and the score is equal or over 50).

Table 2: Equivalence Relationship Between Label and Truthfulness Score

| Label Type | Truthfulness Score Range for Same News |
|---|---|
| *True* | score=100 |
| *Mostly True* | 50<score<100 |
| *Half True* | score=50 |
| *Barely True* | 0<score<50 |
| *False, Pants on Fire* | score=0 |

---

[5]https://www.politifact.com/article/2018/feb/12/principles-truth-o-meter-politifacts-methodology-i/

Table 3: Case Study in Multi-views Credibility Assessment Criteria

| News Statement | Journalists' Fact-Checking Article | ChatGPT's Response |
| --- | --- | --- |
| Clinton said that in 1928, school boards sent kids home with a note warning that if Al Smith were elected president, they would not be allowed to own or read a Bible. | **Label:** *Half True*; It appears that the reference tracks back to a 1932 book, and a newspaper article before that, that spoke of a Sunday school in Daytona Beach passing out a note. There is a chance that other Sunday schools in other states did likewise, but we don't have proof of that. There is a big difference between a Sunday school and a public school. | **Truthfulness Score:** *0/100*; Al Smith was the Democratic nominee for President in 1928, but he was not elected. Furthermore, there is no evidence that school boards sent home letters with children saying that if Al Smith was elected president, they would not be allowed to have or read a Bible. |
| Trump said that "workers in the United States are making less than they were almost 20 years ago, and yet they are working harder." | **Label:** *Barely True*; Adhering strictly to Trump's language, he's the incorrect median, inflation-adjusted wages are higher now than they were for 18 out of the 20 previous years, and hours worked are lower now than in any previous year except for the aftermath of the Great Recession. That said, it doesn't take too many tweaks to Trump's language to make the claim accurate, either by changing the statistic in question or changing the time frame. | **Truthfulness Score:** *80/100*; This statement is largely accurate, as wages for most workers in the United States have stagnated or declined in real terms since the early 2000s. Additionally, many workers are working longer hours and taking on additional jobs to make ends meet. However, there are some exceptions to this trend, such as certain high-skilled professions, which have seen wage increases over the same period. |

The relationship between labels from journalists and truthfulness scores from ChatGPT is shown in Figure 8 (c). We observe a higher level of consistency (i.e., the percentage of *Equivalence* relationship) between human journalists and ChatGPT in assessing the fact-checking results of news on *Social Media* (i.e., 47.72%) compared to news related to two political parties (i.e., 22.96% for Democrat, 26.52% for Republican), which implies there is a greater level of consistency in the news from *Social Media* than politicians. Besides, the percentage of *Hallucination Difference* appearing in *Republican* is the highest, reaching 42.16%. This means that in nearly half of the cases, the fact-checking assessments from human journalists and ChatGPT are opposite, indicating a greater level of controversy in the news from *Republican* than *Democrat*.

Further analysis reveals potential inherent biases in human journalists' perspectives. In Figure 8 (d), we present two cases examining biases from the viewpoints of journalists who primarily focus on news from the two political parties. The upper half of the figure represents the distribution of fact-checking results given by human journalists, while the lower half represents the distribution of results given by ChatGPT. It is evident that these journalists tend to have a negative bias towards political affiliations they do not favor, resulting in stricter evaluations. This phenomenon aligns with previous research findings (Budak et al., 2016) that manual annotations may exhibit inherent biases, manifested as a greater tendency to criticize non-preferred affiliations.

Moreover, in order to intuitively show the difference in specific evaluations of a news article from multi-views (i.e. human journalists and ChatGPT), two representative examples from the ChatGPT-FC dataset are shown in Table 3, which reveals the difference in credibility assessment criteria between journalists and ChatGPT. We can observe that human journalists have inconsistent scoring criteria for statements with factual flaws: the distinction between *Half True* and *Barely True* is ambiguous. In comparison, ChatGPT relies more on factual evidence or statistical data for evaluation, providing a more consistent standard.

Therefore, ChatGPT can offer a corrective perspective on journalists' articles and judgments from an alternative standpoint, which can help address inherent biases present in the content and labels provided by human journalists in ChatGPT-FC. The presence of more factual information leads to fairer judgments. With the assistance of ChatGPT responses in our dataset, fact-checking can be approached from a more objective perspective.

## 6 Conclusion

We present ChatGPT-FC, the first publicly available benchmark dataset for fake news detection that combines human verification and ChatGPT assistance, and conduct a quantitative analysis to compare and contrast the differences between human journalists and LLM in performing fact-checking. Moreover, we propose a framework to analyze the inherent biases caused by political framing from content and labeling perspectives, and we find that the labels provided by human journalists tend to be more ambiguous and the judgments from ChatGPT are more fact-based. These findings highlight the potential of LLM-powered fact-checking systems to enhance the objectivity and reliability of news fact-checking processes.

# Limitations

Although we designed a highly effective and reasonable prompt collection method to gather ChatGPT's fact-checking responses for over 20,000 news articles, there are still a few cases where ChatGPT fails to generate valid responses (i.e., 0.04%), indicating that there are still some flaws in the design of the prompts. We believe that the prompt design in fact-checking can be further improved in future work.

# References

Fatemeh Torabi Asr, Mohammad Mazraeh, Alexandre Lopes, Vasundhara Gautam, Junette Gonzales, Prashanth Rao, and Maite Taboada. 2021. The gender gap tracker: Using natural language processing to measure gender bias in media. *PloS one*, 16(1):e0245533.

Yejin Bang, Samuel Cahyawijaya, Nayeon Lee, Wenliang Dai, Dan Su, Bryan Wilie, Holy Lovenia, Ziwei Ji, Tiezheng Yu, Willy Chung, Quyet V. Do, Yan Xu, and Pascale Fung. 2023. A multitask, multilingual, multimodal evaluation of chatgpt on reasoning, hallucination, and interactivity. *ArXiv*, abs/2302.04023.

Rodrigo Barbado, Oscar Araque, and Carlos A Iglesias. 2019. A framework for fake review detection in online consumer electronics retailers. *Information Processing & Management*, 56(4):1234–1244.

Alberto Barrón-Cedeno, Israa Jaradat, Giovanni Da San Martino, and Preslav Nakov. 2019. Proppy: Organizing the news based on their propagandistic content. *Information Processing & Management*, 56(5):1849–1864.

Ceren Budak, Sharad Goel, and Justin M Rao. 2016. Fair and balanced? quantifying media bias through crowdsourced content analysis. *Public Opinion Quarterly*, 80(S1):250–271.

Joanna M Burkhardt. 2017. History of fake news. *Library Technology Reports*, 53(8):5–9.

Nadia K Conroy, Victoria L Rubin, and Yimin Chen. 2015. Automatic deception detection: Methods for finding fake news. *Proceedings of the association for information science and technology*, 52(1):1–4.

Emmanuel Gbenga Dada, Joseph Stephen Bassi, Haruna Chiroma, Adebayo Olusola Adetunmbi, Opeyemi Emmanuel Ajibuwa, et al. 2019. Machine learning for email spam filtering: review, approaches and open research problems. *Heliyon*, 5(6):e01802.

Jeffrey Dastin. 2018. Amazon scraps secret ai recruiting tool that showed bias against women. In *Ethics of data and analytics*, pages 296–299. Auerbach Publications.

Jacob Devlin, Ming-Wei Chang, Kenton Lee, and Kristina Toutanova. 2018. Bert: Pre-training of deep bidirectional transformers for language understanding. *arXiv preprint arXiv:1810.04805*.

Arianna D'Ulizia, Maria Chiara Caschera, Fernando Ferri, and Patrizia Grifoni. 2021. Fake news detection: a survey of evaluation datasets. *PeerJ Computer Science*, 7:e518.

Luciano Floridi and Massimo Chiriatti. 2020. Gpt-3: Its nature, scope, limits, and consequences. *Minds and Machines*, 30:681–694.

Rebecca Hofstein Grady, Peter H Ditto, and Elizabeth F Loftus. 2021. Nevertheless, partisanship persisted: Fake news warnings help briefly, but bias returns with time. *Cognitive research: principles and implications*, 6:1–16.

Zhijiang Guo, Michael Schlichtkrull, and Andreas Vlachos. 2022. A survey on automated fact-checking. *Transactions of the Association for Computational Linguistics*, 10:178–206.

Cecilia Kang and Adam Goldman. 2016. In washington pizzeria attack, fake news brought real guns. *New York Times*, 5:A1.

Jisu Kim, Jihwan Aum, SangEun Lee, Yeonju Jang, Eunil Park, and Daejin Choi. 2021. Fibvid: Comprehensive fake news diffusion dataset during the covid-19 period. *Telematics and Informatics*, 64:101688.

David MJ Lazer, Matthew A Baum, Yochai Benkler, Adam J Berinsky, Kelly M Greenhill, Filippo Menczer, Miriam J Metzger, Brendan Nyhan, Gordon Pennycook, David Rothschild, et al. 2018. The science of fake news. *Science*, 359(6380):1094–1096.

Stephan Lewandowsky, Michael Jetter, and Ullrich KH Ecker. 2020. Using the president's tweets to understand political diversion in the age of social media. *Nature communications*, 11(1):5764.

Junyi Li, Tianyi Tang, Wayne Xin Zhao, and Ji-Rong Wen. 2021. Pretrained language models for text generation: A survey. *arXiv preprint arXiv:2105.10311*.

OpenAI. 2023. Gpt-4 technical report. *arXiv preprint arXiv:2303.08774*.

Long Ouyang, Jeffrey Wu, Xu Jiang, Diogo Almeida, Carroll Wainwright, Pamela Mishkin, Chong Zhang, Sandhini Agarwal, Katarina Slama, Alex Ray, et al. 2022. Training language models to follow instructions with human feedback. *Advances in Neural Information Processing Systems*, 35:27730–27744.

Olga Papadopoulou, Markos Zampoglou, Symeon Papadopoulos, and Ioannis Kompatsiaris. 2019. A corpus of debunked and verified user-generated videos. *Online information review*, 43(1):72–88.

Parth Patwa, Shivam Sharma, Srinivas Pykl, Vineeth Guptha, Gitanjali Kumari, Md Shad Akhtar, Asif Ekbal, Amitava Das, and Tanmoy Chakraborty. 2021. Fighting an infodemic: Covid-19 fake news dataset. In *Combating Online Hostile Posts in Regional Languages during Emergency Situation: First International Workshop, CONSTRAINT 2021, Collocated with AAAI 2021, Virtual Event, February 8, 2021, Revised Selected Papers 1*, pages 21–29. Springer.

Peng Qi, Yuyan Bu, Juan Cao, Wei Ji, Ruihao Shui, Junbin Xiao, Danding Wang, and Tat-Seng Chua. 2022. Fakesv: A multimodal benchmark with rich social context for fake news detection on short video platforms. *arXiv preprint arXiv:2211.10973*.

Xipeng Qiu, Tianxiang Sun, Yige Xu, Yunfan Shao, Ning Dai, and Xuanjing Huang. 2020. Pre-trained models for natural language processing: A survey. *Science China Technological Sciences*, 63(10):1872–1897.

Prashanth Rao and Maite Taboada. 2021. Gender bias in the news: A scalable topic modelling and visualization framework. *Frontiers in Artificial Intelligence*, 4:664737.

Shaina Raza, Deepak John Reji, and Chen Ding. 2022. Dbias: detecting biases and ensuring fairness in news articles. *International Journal of Data Science and Analytics*, pages 1–21.

Nils Rethmeier and Isabelle Augenstein. 2023. A primer on contrastive pretraining in language processing: Methods, lessons learned, and perspectives. *ACM Computing Surveys*, 55(10):1–17.

Ellen Riloff and Janyce Wiebe. 2003. Learning extraction patterns for subjective expressions. In *Proceedings of the 2003 conference on Empirical methods in natural language processing*, pages 105–112.

Erica Scharrer and Srividya Ramasubramanian. 2015. Intervening in the media's influence on stereotypes of race and ethnicity: The role of media literacy education. *Journal of Social Issues*, 71(1):171–185.

Tal Schuster, Darsh J Shah, Yun Jie Serene Yeo, Daniel Filizzola, Enrico Santus, and Regina Barzilay. 2019. Towards debiasing fact verification models. *arXiv preprint arXiv:1908.05267*.

Kai Shu, Amy Sliva, Suhang Wang, Jiliang Tang, and Huan Liu. 2017. Fake news detection on social media: A data mining perspective. *ACM SIGKDD explorations newsletter*, 19(1):22–36.

Jiasheng Si, Yingjie Zhu, and Deyu Zhou. 2022. Exploring faithful rationale for multi-hop fact verification via salience-aware graph learning. *arXiv preprint arXiv:2212.01060*.

Eugenio Tacchini, Gabriele Ballarin, Marco L Della Vedova, Stefano Moret, and Luca De Alfaro. 2017. Some like it hoax: Automated fake news detection in social networks. *arXiv preprint arXiv:1704.07506*.

Fatemeh Torabi Asr and Maite Taboada. 2019. Big data and quality data for fake news and misinformation detection. *Big Data & Society*, 6(1):2053951719843310.

Sander Van der Linden, Costas Panagopoulos, and Jon Roozenbeek. 2020. You are fake news: Political bias in perceptions of fake news. *Media, Culture & Society*, 42(3):460–470.

Andreas Vlachos and Sebastian Riedel. 2014. Fact checking: Task definition and dataset construction. In *Proceedings of the ACL 2014 workshop on language technologies and computational social science*, pages 18–22.

William Yang Wang. 2017. " liar, liar pants on fire": A new benchmark dataset for fake news detection. *arXiv preprint arXiv:1705.00648*.

Sam Wineburg and Sarah McGrew. 2016. Evaluating information: The cornerstone of civic online reasoning.

Chen Yang, Xinyi Zhou, and Reza Zafarani. 2021. Checked: Chinese covid-19 fake news dataset. *Social Network Analysis and Mining*, 11(1):58.

Jiawei Zhang, Bowen Dong, and S Yu Philip. 2020. Fakedetector: Effective fake news detection with deep diffusive neural network. In *2020 IEEE 36th international conference on data engineering (ICDE)*, pages 1826–1829. IEEE.

## A  Designed Prompt Example

**News Statement:** The earthquake in Turkey is a "scripted" use of geo engineering weather modification HAARP.
**News Creator:** Facebook posts
**Instruction Prompt:** Facebook posts said "The earthquake in Turkey is a "scripted" use of geo engineering weather modification HAARP." Please give the statement a truthfulness score range from 0 to 100 and justify this score.
**ChatGPT Generated Answer:** I would give this statement a truthfulness score of 0 out of 100. This statement is not supported by any scientific evidence and is not based on any facts. There is no evidence to suggest that the earthquake in Turkey was caused by geo engineering weather modification HAARP. This statement is purely speculative and should not be taken seriously.

## B  Generated Invalid Responses by ChatGPT

We identified three categories of ineffective replies as shown in Table 4 after analyzing the responses in ChatGPT where truthfulness scores are not successfully generated. Among all the invalid responses, 40% of the responses are categorized as *Invalid Truthfulness Score*, 40% as the *Invalid Fact-checks* cases, and 20% as *Invalid Answer*.

Table 4: Invalid ChatGPT Answers

| Website fact-checks label | ChatGPT's answer | Wrong type |
|---|---|---|
| False | This statement is false and has no truthfulness score. Iceland has not declared that all religions are mental disorders. This statement is likely an exaggeration of Iceland's recent decision to ban circumcision for non-medical reasons. | Invalid Truthfulness Score |
| Pants on Fire | This statement is not verifiable and therefore cannot be given a truthfulness score. Without any evidence to back up the claim, it is impossible to determine the accuracy of the statement. | Invalid Fact-checks |
| Barely-True | La veracidad de esta declaración se puede calificar con una puntuación entre 80 y 100. Esto se debe a que hay una gran cantidad de informes de fuentes confiables que respaldan esta afirmación. Los precios de la gasolina han aumentado significativamente en todo el país durante el último año, y los informes de precios de la gasolina de la Administración de Información de Energía muestran que los precios promedio de la gasolina han aumentado un 25% desde el año pasado. | Invalid Answer |